\documentclass[11pt,a4paper]{article}
\usepackage{psfrag,graphicx,fancyhdr}
\usepackage{amsmath,times,mathptmx}
\usepackage{subfigure,latexsym,cite}

\usepackage[left, pagewise]{lineno}

\usepackage{graphicx}
\usepackage{mathptmx}
\usepackage{amsmath}
\usepackage[table]{xcolor}
\usepackage{multirow}
\usepackage{changebar}
\usepackage{array}
\usepackage{multirow}
\usepackage{tabularx}
\usepackage{amsfonts,amssymb}
\usepackage{svg}

\newcommand{\mrm}[1]{_\mathrm{#1}}

\usepackage{setspace}

\makeatletter
\renewcommand\section{\@startsection {section}{1}{\z@}%
                                   {-3.5ex \@plus -1ex \@minus -.2ex}%
                                   {2.3ex \@plus.2ex}%
                                   {\normalfont\normalsize\scshape\bfseries\uppercase}}
\renewcommand\subsection{\@startsection{subsection}{2}{\z@}%
                                     {-3.25ex\@plus -1ex \@minus -.2ex}%
                                     {1.5ex \@plus .2ex}%
                                     {\normalfont\normalsize\bfseries}}
\renewcommand\subsubsection{\@startsection{subsubsection}{3}{\z@}%
                                     {-3.25ex\@plus -1ex \@minus -.2ex}%
                                     {1.5ex \@plus .2ex}%
                                     {\normalfont\normalsize\itshape}}
\long\def\@makecaption#1#2{%
  \vskip\abovecaptionskip
  \sbox\@tempboxa{\textbf{\MakeUppercase{#1} #2}}%
  \ifdim \wd\@tempboxa >\hsize
    \textbf{\MakeUppercase{#1} #2}\par
  \else
    \global \@minipagefalse
    \hb@xt@\hsize{\hfil\box\@tempboxa\hfil}%
  \fi
  \vskip\belowcaptionskip}
\def\@cite#1#2{({\textit{#1}}\if@tempswa , #2\fi)}
\makeatother

\allowdisplaybreaks

\begin{document}

\vspace{3cm}
\noindent
\textbf{\large A Bidirectional Long Short Term Memory Approach for Infrastructure Health Monitoring Using On-board Vibration Response}
\\[2ex]
Authors:
\\[2ex]
Reza Riahi Samani\\Delft Center for Systems and Control\\Delft University of
Technology\\E-mail: r.riahisamani@tudelft.nl
\\[2ex]
Alfredo Nunez Vicencio\\Section of Railway Engineering\\Delft University of Technology\\E-mail: a.a.nunezvicencio@tudelft.nl
\\[2ex]
Bart De Schutter\\Delft Center for Systems and Control\\Delft University
of Technology\\
E-mail: b.deschutter@tudelft.nl

\newpage

\noindent
\textbf{ABSTRACT}\\
   The growing volume of available infrastructural monitoring data enables the development of powerful data-driven approaches to estimate infrastructure health conditions using direct measurements. This paper proposes a deep learning methodology to estimate infrastructure physical parameters, such as railway track stiffness, using drive-by vibration response signals. The proposed method employs a Long Short-term Memory (LSTM) feature extractor accounting for temporal dependencies in the feature extraction phase, and a bidirectional Long Short-term Memory (BiLSTM) networks to leverage bidirectional temporal dependencies in both the forward and backward paths of the drive-by vibration response in condition estimation phase. Additionally, a framing approach is employed to enhance the resolution of the monitoring task to the beam level by segmenting the vibration signal into frames equal to the distance between individual beams, centering the frames over the beam nodes. The proposed LSTM-BiLSTM model offers a versatile tool for various bridge and railway infrastructure conditions monitoring using direct drive-by vibration response measurements. The results demonstrate the potential of incorporating temporal analysis in the feature extraction phase and emphasize the pivotal role of bidirectional temporal information in infrastructure health condition estimation. The proposed methodology can accurately and automatically estimate railway track stiffness and identify local stiffness reductions in the presence of noise using drive-by measurements. An illustrative case study of vehicle-track interaction simulation is used to demonstrate the performance of the proposed model, achieving a maximum mean absolute percentage error of 1.7\% and 0.7\% in estimating railpad and ballast stiffness, respectively.

\noindent\textbf{Keywords}: Deep Learning, Infrastructure Health Monitoring, Vibration Response, Railway Track Stiffness
   
\newpage

\section{Introduction}


Infrastructure health monitoring is a fundamental component of maintaining a safe and reliable infrastructure network. Continuous monitoring over time and across the infrastructure network facilitates the infrastructure health conditions estimation, which can be used for condition-based maintenance strategies \cite{ghofraniBayesian2019, mohammadiExploring2019a}. This process involves assessing infrastructure damage and tracking changes in physical properties, such as modal frequencies, modal mass, modal damping, stiffness, and mode shapes  \cite{radzienskiImprovementDamageDetection2011, neuholdMeasurement2020,movagharIntelligent2020, movagharBayesian2022}.

In railway systems, monitoring the physical properties of infrastructure is crucial for capturing changes due to dynamic train loading and aging track components. Track stiffness forms a primary feature determining the degradation progress of railway infrastructure \cite{nielsenDegradationRailwayTrack2020, hoelzlVoldKalmanFilter2023}. A sudden change in track stiffness is often related to changes in track type or associated to local damage to the ballast, substructure, and subgrade (fouled ballast, mud pumping, or hanging ties). Moreover, track stiffness is influenced by the superstructure and missing/degraded fastening components \cite{shenEvaluatingRailwayTrack2023a}. Given the determining nature of track stiffness for railway condition estimation, accurate measurements of this property can significantly support maintenance decisions.

Infrastructure health conditions can be measured indirectly under both loaded and unloaded conditions using dynamic loads or external excitation forces. Two common techniques for measuring health conditions under loaded conditions are pass-by and vehicle-based (drive-by) measurements \cite{fernandez-bobadillaModernTendenciesVehicleBased2023}. Pass-by measurements are typically used for specific locations of interest and involve placing a network of accelerometers to record vibration responses at strategic points, such as bridges. These methods incur costs for sensor deployment and limit measurements to a single location. Consequently, vehicle-based measurements have become increasingly preferred since they present high-accuracy and cost-effective measurements along the entire infrastructure network.

Traditionally, infrastructure physical parameter estimation methods relied on sophisticated identification algorithms to solve inverse problems in modal parameter estimation \cite{alvandiAssessmentVibrationbasedDamage2006, quirkeDrivebyDetectionRailway2017, zhuIdentificationRailwayBallasted2018}. For example, \cite{quirkeDrivebyDetectionRailway2017, zhuIdentificationRailwayBallasted2018} investigated track stiffness estimation using optimization techniques namely, cross-entropy \cite{quirkeDrivebyDetectionRailway2017} and adaptive regularization \cite{zhuIdentificationRailwayBallasted2018} to solve the identification optimization problem and to infer the railway track stiffness profile.  However, implementing such algorithms is often computationally expensive and time-consuming \cite{avciReviewVibrationbasedDamage2021}. Furthermore, the centralized nature of parametric damage detection methods makes them infeasible for real-time damage detection applications, especially as infrastructure data continues to grow exponentially. Hence, recent advances in machine learning, particularly deep learning techniques, offer promising alternatives for overcoming these challenges \cite{phusakulkajornArtificialIntelligenceRailway2023a, zamanVideo2018, rafeImputing2024,rafeExploring2024a, sonyMulticlass2021, hungStructural2020}.

In the current paper, we further explore the estimation of infrastructure physical parameters through deep learning approaches. Railway inspection often relies on Axle Box Acceleration (ABA) monitoring systems \cite{shenEvaluatingRailwayTrack2023a, yanDeveloping2023, chellaswamyOptimized2021}. Hence, we develop a novel deep learning framework to estimate infrastructure health parameters using ABA measurements and demonstrate the application of our model utilizing simulated ABA signals for different track stiffness parameters. Furthermore, our methodology holds promise for broader applications in infrastructure inspections utilizing drive-by vibration measurements. For further information on drive-by infrastructure inspections, readers can refer to \cite{hajializadehDeepLearningbasedIndirect2023}.



\subsection{Deep learning in structure health monitoring using vibration response}

 Deep learning techniques have increasingly gained attention in infrastructure health monitoring in recent years. The strength of these techniques to detect and locate damage directly from vibration response without any need for data preprocessing or hand-crafted feature extraction is highly promising compared to traditional vibration-based structure monitoring techniques \cite{avciReviewVibrationbasedDamage2021}. Deep learning techniques generally require a two-stage process including a damage-sensitive feature extraction through the raw acceleration signals, and processing the extracted features to assess the health state of the structure \cite{avciReviewVibrationbasedDamage2021, pathirageStructuralDamageIdentification2018a}. Accordingly, an autoencoder-based framework has been proposed by \cite{pathirageStructuralDamageIdentification2018a} for structural damage identification, comprising two main components: dimensionality reduction and relationship learning. The first component reduces the dimensionality of the original input vector, and the second component learns the relationship between the features and the stiffness reduction. Furthermore, \cite{huangQuantificationDynamicTrack2022} proposed a convolutional neural network (CNN) framework to predict the track dynamic stiffness using the ABA measurements in real-time, including a comparison of the performance between the standard CNN and dilated CNN algorithms. Both models performed well considering their accuracy, with the dilated CNN requiring less computation time in both the training and deployment processes. While CNN effectively captures relevant information within a neighborhood of samples, it often struggles to learn long-term dependencies in sequential datasets, which is relevant for railway track parameter estimation over a long period of data. To address this, \cite{le-xuanNovelApproachModel2024} proposed a 1DCNN-LSTM-ResNet architecture to identify structural damages based on time-dependent data. The model employs one-dimensional CNN (1DCNN) for feature extraction, Long Short-term Memory (LSTM) for recognizing long-term dependencies, and ResNet to counteract the vanishing gradient problem during deep network training. The proposed architecture outperformed 1DCNN, LSTM, and their combination in diagnosing the damage states of the Z24 bridge located in the Bern district near Solothurn, Switzerland. In \cite{lockeUsing2020}, the frequency spectrum of simulated acceleration signals was utilized to develop and train a 1DCNN deep learning algorithm. The paper \cite{lockeUsing2020} showed that a one-dimensional variant of VGG convolutional networks can detect bridge damage across a range of real-world noise signals. In fact, convolutional neural networks (CNNs) are designed to efficiently extract spatial and hierarchical features from data, making them particularly effective for inputs like vibration response signals. The process of applying filters and pooling in a CNN helps capture essential features while reducing spatial dimensions, which not only enhances computational efficiency but also preserves critical information \cite{lockeUsing2020}.

Moreover, structure health monitoring has been addressed with anomaly detection approaches. The papers \cite{jieTrackVibrationSequence2023a}, \cite{sharmaRealtimeStructuralDamage2023} developed LSTM-based anomaly detection frameworks to identify vibration sequence anomalies in subway tracks and bridges, respectively. The LSTM model was trained only on normal sequences, and the anomaly score was estimated via the reconstruction error between the model input sequence and output sequence. The input sequence was identified as an anomaly sequence through comparison to the anomaly threshold value. Although the anomaly detection approach is highly applicable when there are limited available data of malfunction conditions, since the model's training process is biased towards normal cases, it may not generalize well to abnormal situations.

\subsection{Contributions of the paper}

In this paper, we propose an innovative model architecture, namely, LSTM-BiLSTM networks, to estimate infrastructure physical parameters using vehicle-based vibration responses. This work makes the following contributions with respect to the state-of-the-art. First, our work highlights the pivotal role of the bidirectional temporal dependencies in both the forward and backward paths of the vibration response of the infrastructure. The idea stems from the fact each sequence of vibration responses is influenced by both the forward and backward paths of the infrastructure's vibration response. By using BiLSTM networks, we can effectively capture and integrate these temporal relations, leading to more accurate infrastructural health condition estimations. Second, we enhance the infrastructure monitoring resolution to beam nodes using the proposed framing approach to leverage accurate vibration signal positions. This framing approach segmentizes the vibration response within bearing spans over the beam nodes, which facilitates localized condition estimation at individual beam resolutions. Thirdly, the methodology employs a novel feature extraction method based on an LSTM layer and proposes the potential of incorporating temporal relations even at the feature extraction level. Our model exhibits excellent performance in a case study focusing on railway track segments, demonstrating accurate estimation of railpad and ballast stiffness, and identifying local stiffness reductions in a noisy environment.


\section{Preliminaries}
\label{sec_IV}

\subsection{LSTM cell}

\begin{figure}[thpb]
      \centering
      \makebox{\parbox{3.3in}{%
      \centering\includegraphics[scale=0.60]{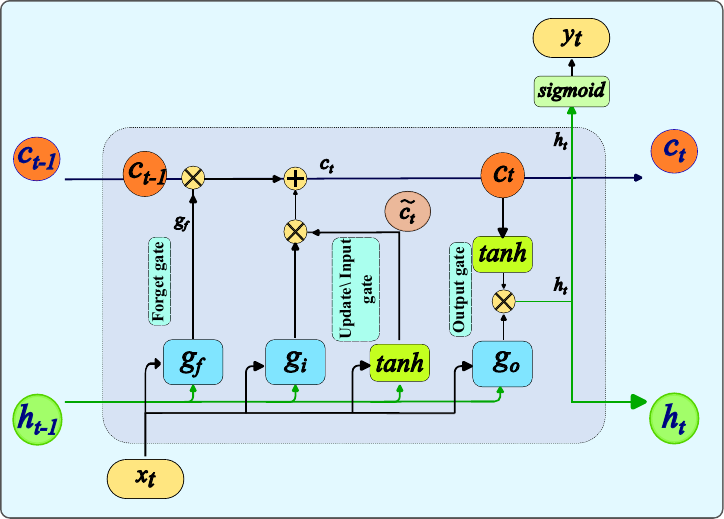}
      }}   
      \caption{Long Short Term Memory (LSTM) cell. The fundamental components
of an LSTM cell are a forget, input, and output gate, updating cell state, and hidden state.}
      \label{fg:lstm}
   \end{figure}

\medskip LSTM was introduced as an enhancement over traditional Recurrent Neural Networks (RNNs) for long-term dependency \cite{hochreiterLongShortTermMemory1997}. The LSTM network captures temporal dependencies with additional memory cells. An LSTM memory cell controls the flow of information and handles the memory efficiently using three types of gates: input, forget, and output gate as shown in Figure \ref{fg:lstm}. Given a time series input $x= (x_1, \dots, x\mrm{T})$, where $T$ represents the last time step $t$, LSTM updates the hidden state $h_t$ using current input information $x_t$, hidden prior information state $h_{t-1}$, input gate $g_{\mathrm{i},t}$, forget gate $g_{\mathrm{f},t}$, output gate $g_{\mathrm{o},t}$ and a memory cell $c_t$. The mathematical expression for a forward pass of an LSTM unit can be defined as

\begin{flalign}
    &g_{\mathrm{i},t}  = \sigma(W_{x\mathrm{i}} x_{t} + W_{h\mathrm{i}}h_{t-1} + b\mrm{i})
    \label{}& \\[1mm]
    &g_{\mathrm{f},t} = \sigma(W_{x\mathrm{f}} x_{t} + W_{h\mathrm{f}}h_{t-1} + b\mrm{f})
    \label{}& \\[1mm]
    &g_{\mathrm{o},t}  = \sigma(W_{x\mathrm{o}} x_{t} + W_{h\mathrm{o}}h_{t-1} + b\mrm{o})
    \label{}&\\[1mm]
    &\tilde{c}_{t}  = \tanh(W_{x\mathrm{\tilde{c}}}x_{t} + W_{h\mathrm{\tilde{c}}}h_{t-1} + b\mrm{\tilde{c}})
    \label{}&\\[1mm]
    &c_{t} = g_{\mathrm{i},t} \otimes \tilde{c}_{t} + g_{\mathrm{f},t}\otimes c_{t-1}   
    \label{}&\\[1mm]
    &h_{t} = g_{\mathrm{o},t} \otimes \tanh(c_{t})  
    \label{eq:hidouttt}&\\[1mm]
    &y_{t} = \sigma(W_yh_{t} + b_{y})&
\end{flalign}

\noindent where the notations $\otimes$ is an operator for point-wise multiplication of vectors, $\varphi$, and $\sigma$ are the $\tanh$ and $\mathrm{sigmoid}$ activation functions respectively, and $W$ and $b$ are respectively weight matrices and bias parameters for each LSTM cell. Moreover, $\tilde{c}_{t}$ denotes the cell input activation vector, and $c_{t}$ and $c_{t-1}$ denotes the cell state values at time steps $t$ and $t-1$ respectively. The input gate $g_{\mathrm{i},t}$ determines the cell state that needs to be updated, the forget gate $g_{\mathrm{f},t}$ decides what information should be overlooked, and the output gate $g_{\mathrm{o},t}$ determines which part of the cell state should be exported.

A cell state $c_{t}$ consists of two components: i) $g_{\mathrm{i},t} \otimes \tilde{c}_{t}$, which includes the relevant information from $\tilde{c}_{t}$ through $g_{\mathrm{i},t}$ and ii) $g_{\mathrm{f},t}\otimes c_{t-1}$, which forgets the less-relevant information from $c_{t-1}$. In the final step, the hidden state $h_t$  is employed to predict $y_t$ at each time step.  The entire network is parameterized with the weight matrices and bias parameters and is learned during backpropagation.

\subsection{1D Convolutional Neural Networks}

One-dimension Convolutional Neural Networks (1DCNNs) are recognized in deep learning for their proficiency in handling local relationships in sequential and time-series data. A basic 1DCNNs comprises several 1D filters in the 1DCNN layer and 1D pooling layers to capture the spatial patterns by applying the convolution operation. During this process, filters slide across time steps, creating feature maps that capture detailed and abstract representations of the data's patterns. Feature maps are essentially the outputs of the convolution process, depicting different aspects of the input data. Moreover, pooling layers complement this by downsampling the outcome of the convolution process, effectively emphasizing important features while reducing computational load. Max pooling or average pooling layers are commonly used to summarize the feature maps by reducing their spatial dimensions and highlighting salient features in the process.

The formula of one typical convolutional layer is:

\begin{flalign}
&h_f = \mathrm{conv1D}(W_{f},x) + b_f
    \label{eq:hidout}&
\end{flalign}

\noindent where $h_f$, $W_{f}$ and $b_f$ are respectively the output vector, weight matrices, and bias parameter of the filter $f$, while $x$ is the input vector and $\mathrm{conv1D}$ is the 1D convolution operator whose $i^{th}$ output is calculated by the following formula:

\begin{flalign}
&\mathrm{conv1D}(W_f,x(i)) =  W_{f} \otimes x (i) = \sum_{j=1}^{N_f} w_{fj} \; x_{i-j}
    \label{eq:hidoutt}&
\end{flalign}

\noindent where $N_f$ is the length of the filter $f$, and $W_{fj}$ is the $j^{th}$ element of matrices $W_f$.


\section{Methodology}

\subsection{Problem statement}
Infrastructure health condition estimating using vehicle-based vibration response requires spatial and temporal analysis of the vibration signals. We approach the drive-by vibration response as a sequence of vibration signals associated with the sequence of beams and their physical properties. Our framing approach segmentises the vibration response signal with the spans, equal to the distances between individual beams, and located at the beam nodes. These bearing spans are associated with the lengths where, under normal conditions, a beam node bears a relative load between the beam nodes. By segmenting the vibration signal over the bearing spans at the beam nodes, we leverage domain knowledge about the position of beams and the vibration signal, enhancing the resolution of infrastructure health monitoring at the beam level. For instance, in the case of railway track monitoring, the vibration signal is framed for each sleeper as a beam node, with the length of the bearing span equal to the distance between individual sleepers, centred at each sleeper position. Consequently, the sequential model can estimate health conditions in a sequential manner at the resolution of individual sleepers.

The proposed approach involves a multi-level sequential data analysis. First, the lower-level network performs the feature extraction task over the vibration response segments. Second, the upper-level BiLSTM network takes the extracted features to estimate the health conditions over the sequence of beams. The two steps are performed in sequence with an end-to-end approach. This eliminates defining handcrafted features, which may require expert knowledge and limits a fully automated analysis. The hidden layers of the deep learning network are learned through the backpropagation process corresponding to the last layer of the network. The initial layers may extract more meaningful features, like the sharp changes in the input signal, and the deeper layers may extract more abstract and high-level features. 
Section 3.2 below discusses the lower-level bearing span feature extraction methods, followed by Section 3.3, which discusses the upper-level BiLSTM networks for the estimation of health conditions.

\subsection{Lower level bearing span feature extractor}

\subsubsection{1DCNN}

The idea behind using 1DCNN in the feature extraction phase is to leverage the strength of 1DCNN in capturing local relationships in this phase. The 1DCC can provide features for the upper-level model based on local relations within the input signal. The 1D-CNN architecture used in this paper consists of convolutional and max-pooling layers, followed by a flattening layer and a fully connected layer, which forms the output layer of the feature extraction process. The hyperparameters of the feature extraction network include the number of convolutional layers, the number of filters per layer, filter lengths, and the size of the dense layer, which represents the latent space of the feature extractor.

To optimize the model architecture, grid search is employed for hyperparameter tuning. The resulting network processes high-frequency time-domain signals of length 8,660 and applies three convolutional and max-pooling layers of 32, 64, and 128 filters with a filter length of 8. The output of convolutional layers is activated by a ReLU function to increase the nonlinear expression of networks. In max-pooling layers, the pool size is $1 \times 2$ and the stride is two. The CNNs output is flattened and passed through a dense layer with 128 units to generate the final feature vector for the subsequent health condition estimator.



While 1DCNNs emphasise the local patterns in the input sequence, they can be less efficient in capturing temporal dependencies, especially when the filter size is not sufficiently large. However, Recurrent architectures, such as LSTMs, effectively address this challenge by capturing temporal information using their cell memory. Hence, we also investigate the performance of an LSTM feature extractor in the feature extraction phase.

\subsubsection{LSTM}
In contrast to the previous approach, The LSTM feature extractor captures temporal relationships within the input sequence. While the vibration signal exhibits long-term temporal relations in the upper-level
beam-to-beam context, temporal relations can be captured during the feature extraction phase using an LSTM feature extraction layer. The feature extraction network consists of a single LSTM layer, which processes frames of vibration signals and outputs the hidden state of the last time step as the extracted feature for each frame. This hidden state serves as the feature vector for the subsequent BiLSTM health condition estimator networks. The hyperparameters of the LSTM feature extractor networks include the number of LSTM layers and LSTM units per layer. Grid search tuning identified an optimal architecture with one LSTM layer containing 128 LSTM units.

\begin{figure}[thpb]
      \centering
      \makebox{\parbox{5in}{%
      \includegraphics[scale=1.1]{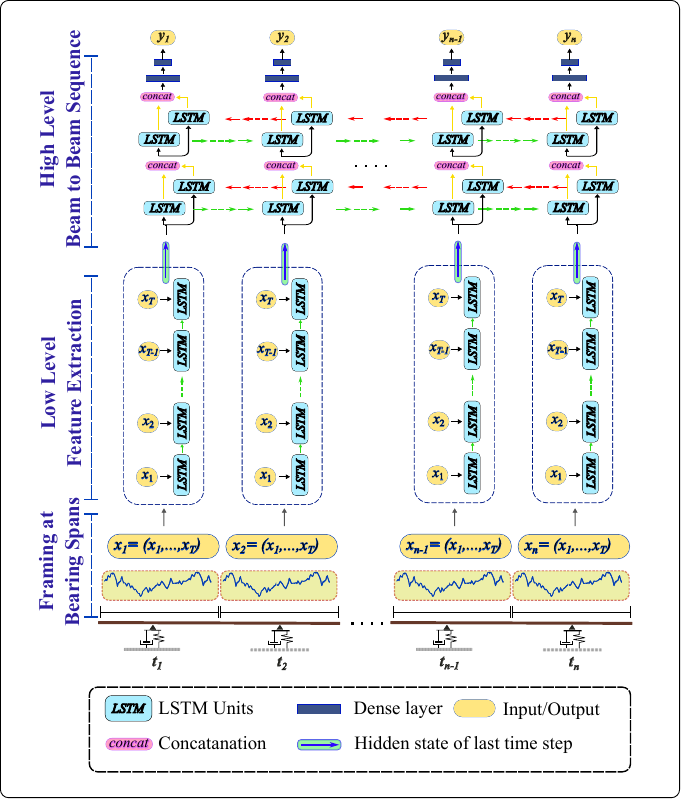}
      }}   
      \caption{Model architecture of the proposed LSTM-BiLSTM model.}
      \label{1dcnn-bi-lstm}
   \end{figure}

\subsection{Upper-level BiLSTM Networks}

After extracting features from the bearing span frames in the lower-level analysis, the output is fed into BiLSTM networks at the upper level to capture the long-term temporal relations between sequences of beams. This approach is based on the physics of vehicle-based vibration response signals, where sequences are related to both the forward and backward paths of the sequence. For example, when there is a defect in one of the beams in a sequence, it is reflected not only in the forward path of the drive-by vibration signal but also in the backward path. Therefore, we propose using BiLSTM networks consisting of two stacked BiLSTM layers with 128 and 64 units, respectively. The outputs of these layers are concatenated and passed to the subsequent layers. The bidirectional information is then fed into a fully connected network with two dense layers consisting of  64 and 2 units, respectively, to produce the final output vector. The key hyperparameters include the number of BiLSTM layers, the units in each layer, and the number of dense layers and the units in each layer. A schematic view of the LSTM-BiLSTM model architecture is provided in Figure \ref{1dcnn-bi-lstm}.

\medskip
\section{Illustrative case study of Railway track  monitoring}
Railway track stiffness is a key indicator of track health and is widely used to monitor infrastructure conditions. It is influenced by various parameters associated with track components, including the rail, sleepers, fastening systems, ballast, and track substructure. Among these, ballast stiffness primarily determines overall track stiffness; however, fastening systems—comprising bolts, clamps, and rail pads—also play a significant role. In this study, track stiffness is characterized using railpad and ballast stiffness parameters, represented as the vector $k = [k_\mathrm{p}, k_\mathrm{b}]^\intercal$. The influence of fastening components, such as bolts and clamps, is considered through the rail pad stiffness parameter. The primary objective of this analysis is to estimate track stiffness parameters based on observed Axle-Box Acceleration (ABA) vibration responses.

\subsection{Data}

\begin{figure}[thpb]
      \centering
      \framebox{\parbox{\linewidth}{%
      \includegraphics[scale=0.85]{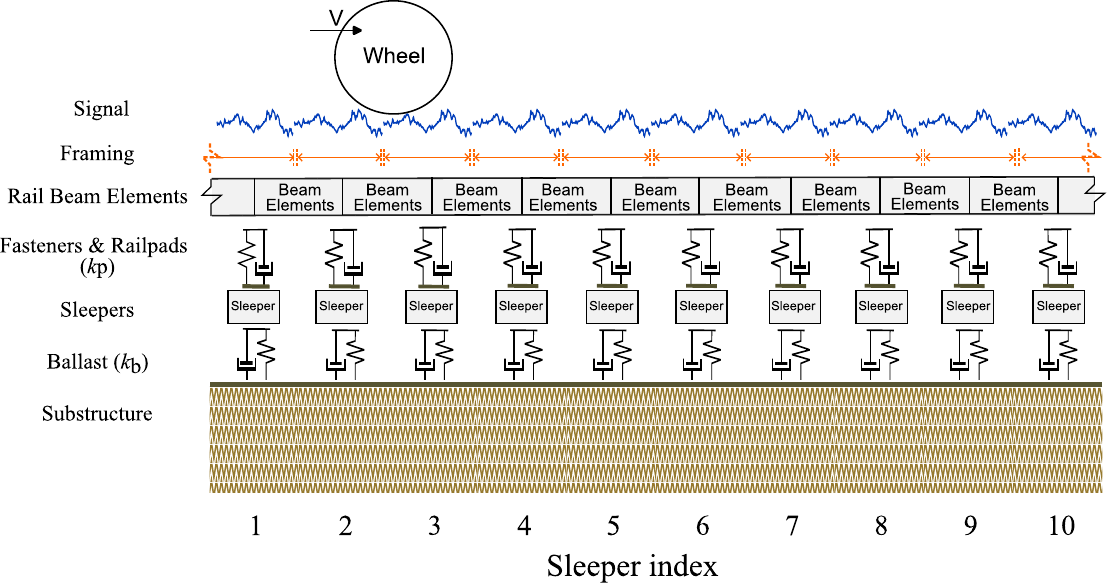}}}
         
      \caption{The layout of the 10 sleepers track segment, adapted from \cite{shenEvaluatingRailwayTrack2023a}.}
      \label{fg:layout}
   \end{figure}

Data used in the illustrative case study is based on the finite-element simulation model proposed in \cite{shenEvaluatingRailwayTrack2023a} and is available online \cite{Vehicle}. In this model, the rail and sleepers are meshed using Timoshenko beam elements, while the ballast and railpads are represented as discrete spring-damper pairs. Clamps and bolts are not explicitly modeled; instead, their stiffness is incorporated into the railpad stiffness, a widely accepted simplification in railway track modeling \cite{shenFast2021, shenEvaluatingRailwayTrack2023a, kaewunruenField2007}. The wheel is simplified as a rigid mass, and the wheel-rail contact is modeled using a Hertzian spring. In real-world railway networks, different track types exist due to variations in rail types, fastening systems, sleeper types, and sleeper spacing, which the current simulation may be limited to account for \cite{shenEvaluatingRailwayTrack2023a}. Figure \ref{fg:layout} illustrates the schematic view of the finite-element model for a 10-sleeper track segment alongside the corresponding framing approach for the vibration response signal. We use this model to simulate ABA measurements at an operational speed of 65 km/h by considering floating stiffness reduction and additive white Gaussian noise scenarios.

The dataset includes cases of both constant track stiffness and local variations in track stiffness, due to changes in railpad and ballast stiffness. The constant track stiffness scenario represents normal railway track conditions, where stiffness remains uniform along the track. To simulate this scenario, ballast and railpad stiffness values are randomly sampled from the stiffness range set R1 in Table 1, following a uniform distribution. The stiffness reduction is considered in two scenarios: reduction in one sleeper and reduction in three sleepers within a 10-sleeper segment at random locations. The stiffness reduction in one sleeper simulates a defect such as a hanging sleeper, where an unsupported sleeper hangs within the railway segment, causing a local reduction in track stiffness at that sleeper. The stiffness reduction in three sleepers simulates a scenario where a reduction in substructure support over a wider length leads to reduced stiffness in multiple sleepers \cite{shenEvaluatingRailwayTrack2023a}. The stiffness of the railpad and ballast are simultaneously reduced to the stiffness range set R2 in Table \ref{table_range}, using a random variable with uniform distribution. The stiffness range set R2 represents multiple faults with varying severities, including degraded or missing railroads or local reductions in ballast stiffness \cite{shenEvaluatingRailwayTrack2023a}. The lower bound of the range set R2 indicates the more harsh scenarios of the component degradation such as ballast crushing or hanging sleepers \cite{shiCritical2023}. The random locations of stiffness reduction within the track segment prevent biases in the spatial distribution of defects, making the simulations more representative of real-world scenarios.


\begin{table}
\centering
\renewcommand{\arraystretch}{1.1}
  \caption{\,\, Range of parameter values used for railpad and ballast stiffness.}
  \label{table_range}
  \begin{tabularx}{0.95\textwidth} { 
      | >{\raggedright\arraybackslash}p{75mm}
      | >{\centering\arraybackslash}X
      | >{\centering\arraybackslash}X | }
   \hline
  Range sets (range of values of the model's parameters)  &  $k_\mathrm{p}\, (\mathrm{N/M})$  & $k_\mathrm{b}\, (\mathrm{N/M})$ \\
  \hline
  R1&$1.5\cdot 10^8$\,-\,$3\cdot10^8$&  $1.6\cdot10^7$\,-\,$2.2\cdot10^7$\\
  R2&$0.1\cdot10^8$\,-\,$1.5\cdot10^8$& $0.4\cdot10^7$\,-\,$1.6\cdot10^7$\\

  \hline    
  \end{tabularx}
  
\end{table}


Noise is an inherent aspect of signal measurement, present even in data acquired from highly accurate sensors. Drive-by vibration responses are also affected by measurement noise due to environmental and vehicle-induced noise. Therefore, an additive white Gaussian noise \(\sim N(0, \sigma^2) \) is added to the ABA signal, where $\sigma$  is the standard deviation of the measurement noise. A noise-to-signal ratio of 15\% is used, with a variance equal to 15\% of the ABA signal power. The additive Gaussian noise in the ABA signal is considered to simulate real-world measurement conditions and evaluate the performance of the proposed model under such conditions \cite{gonzalezEffective2023, malekjafarianMachine2019}. Each of the noise-free and noise-added scenarios consists of 15\,000 records of 10-sleeper track segments, and the datasets are split into training, validation, and test sets, with 60\%, 20\%, and 20\% of the total records, respectively. 

 \medskip

\subsection{Results}

In this section, we present the performance of the proposed LSTM-BiLSTM networks in estimating track stiffness using railway ABA measurements. For comparison, we also consider three other model architectures: 1DCNN-LSTM, LSTM-LSTM, and 1DCNN-BiLSTM. 

\begin{figure}[htp]
      \centering
      \framebox{\parbox{\linewidth}{%
      \includegraphics[scale=0.85, trim=0 360 10 250, clip]{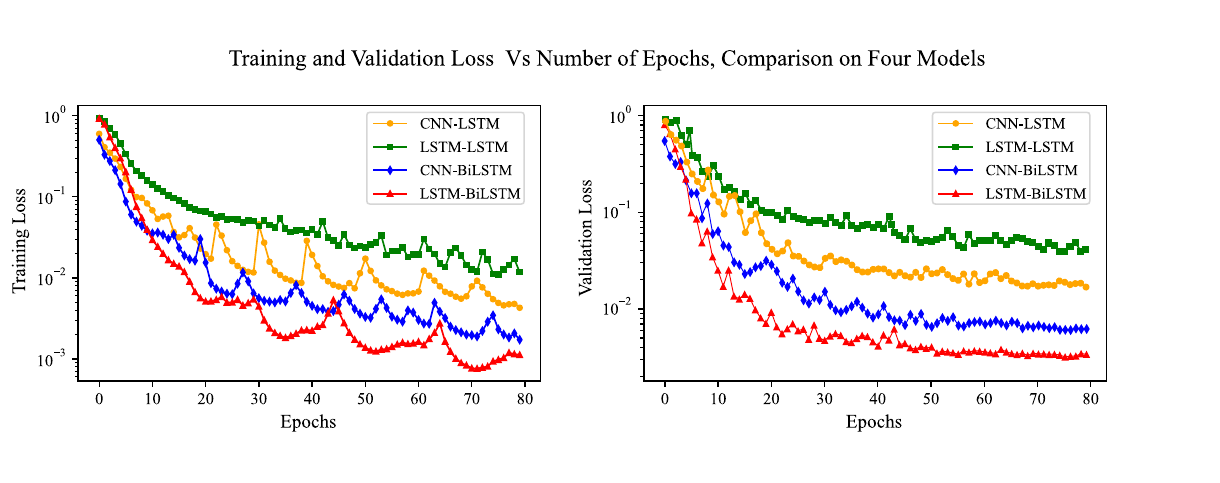}
      }}   
      \caption{Training and validation loss convergence for the noisy dataset.}
      \label{epochs}
   \end{figure}

During the training and validation process, the data is split into training and validation sets. Figure \ref{epochs} shows the training and validation loss convergence of the four models (CNN-LSTM, LSTM-LSTM, CNN-BiLSTM, and LSTM-BiLSTM) on the noise-added datasets, demonstrating the loss convergence versus the number of epochs on a logarithmic scale. The figure indicates that the LSTM-BiLSTM architecture achieves the lowest training and validation loss compared to the other models. Furthermore, the figure shows no evidence of overfitting in the training process. The validation set is used to fine-tune the hyperparameters of the models using a grid search method. The performance of the selected fine-tuned models is then evaluated on the test set, with the results presented in Table \ref{tab:results}. Two evaluation metrics, Mean Absolute Percentage Error (MAPE) and Root Mean Square Error (RMSE), are used to compare the accuracy of the models in stiffness estimation. MAPE measures the average absolute percentage difference between predicted and actual values, while RMSE measures the square root of the average squared differences between predicted and actual values. 

Table \ref{tab:results} shows the estimation accuracy performance of the four models (CNN-LSTM, LSTM-LSTM, CNN-BiLSTM, and LSTM-BiLSTM) on the test set under two conditions: noise-free and noise-added datasets. The performance metrics include Root Mean Squared Error (RMSE) and Mean Absolute Percentage Error (MAPE) for two parameters, railpad ($k\mrm{p}$) and ballast ($k\mrm{b}$) stiffness, along with an overall MAPE.

 In a noise-free environment, the LSTM-BiLSTM model outperforms the other models, achieving the lowest RMSE for $k\mrm{p}$ at 0.82\,(MN/m) and for $k\mrm{b}$ at 0.06\,(MN/m). The MAPE values for LSTM-BiLSTM are also the lowest, with 0.61\% for $k\mrm{p}$ and 0.35\% for $k\mrm{b}$, resulting in an overall MAPE of 0.47\%. The CNN-BiLSTM model demonstrated the second-best performance in the noise-free condition. It recorded an RMSE of 2.81\,(MN/m) for  $k\mrm{p}$ and 0.14\,(MN/m) for $k\mrm{b}$. Its MAPE values are 1.76\% for kp and 0.74\% for $k\mrm{b}$, leading to an overall MAPE of 1.25\%. The CNN-LSTM and LSTM-LSTM models rank below their BiLSTM counterparts, with overall MAPE of 1.66\% and 1.92\% respectively.



Under the 15\% noise condition, the RMSE and MAPE values for all models increase. However, the LSTM-BiLSTM model remains the best performer, with an RMSE of 2.72\,(MN/m) for $k\mrm{p}$ and 0.13\,(MN/m)for $k\mrm{b}$. The MAPE values are 1.70\% for $k\mrm{p}$ and 0.70\% for $k\mrm{b}$, resulting in an overall MAPE of 1.20\%. The CNN-BiLSTM model continued to show the second performance with an RMSE of 6.39\,(MN/m) for $k\mrm{p}$ and 0.38\,(MN/m) for $k\mrm{b}$. The MAPE values were 3.90\% for $k\mrm{p}$ and 1.57\% for $k\mrm{b}$, leading to an overall MAPE of 2.77\%.

 The LSTM-BiLSTM and CNN-BiLSTM models consistently outperform the other two models across both datasets. The LSTM-BiLSTM model, in particular, shows low RMSE and MAPE values, highlighting its superior capability in handling both noise-free and noisy data conditions. The LSTM-LSTM and CNN-LSTM models show almost similar performance in noise-free conditions; however, the LSTM-LSTM shows poorer performance under noise-added conditions. Overall, all models perform better in predicting ballast stiffness than railpad stiffness.

\begin{table}[h]
\caption{Models estimation accuracy performance on the test set in noise-free and (15\%) noisy datasets.}
\label{tab:results}
\renewcommand{\arraystretch}{1.6}
\small
    \begin{center}
     \begin{tabularx}{\textwidth}{| >{\raggedleft\arraybackslash}p{12mm}
      | >{\raggedleft\arraybackslash}X 
      | >{\raggedleft\arraybackslash}X
      | >{\centering\arraybackslash}X
      | >{\centering\arraybackslash}X
      | >{\centering\arraybackslash}X
      | >{\centering\arraybackslash}X 
      | >{\centering\arraybackslash}X
      | >{\centering\arraybackslash}X
      | >{\centering\arraybackslash}X 
      | >{\centering\arraybackslash}X
      | >{\centering\arraybackslash}X
      | >{\centering\arraybackslash}X
      | >{\centering\arraybackslash}X|}
    \hline
    
    \multirow{3}{*}{\textbf{Models}} &  \multicolumn{5}{c|}{\cellcolor[rgb]{0.67,0.93,1.00}Noise-Free Dataset} & \multicolumn{5}{c|}{\cellcolor[rgb]{0.67,0.93,1.00}Noisy Dataset}  \\
    \cline{2-11}
    & \multicolumn{2}{c|}{\cellcolor[gray]{0.9}$k\mrm{p}$} & \multicolumn{2}{c|}{\cellcolor[gray]{0.9}$k\mrm{b}$} & \multicolumn{1}{c|}{\cellcolor[gray]{0.9}\textbf{Overall}} & \multicolumn{2}{c|}{\cellcolor[gray]{0.9}$k\mrm{p}$} & \multicolumn{2}{c|}{\cellcolor[gray]{0.9}$k\mrm{b}$} & \multicolumn{1}{c|}{\cellcolor[gray]{0.9}\textbf{Overall}}\\
    \cline{2-11}
    & \textbf{RMSE} & \textbf{MAPE} & \textbf{RMSE} & \textbf{MAPE}  & \textbf{MAPE }& \textbf{RMSE} & \textbf{MAPE} & \textbf{RMSE} & \textbf{MAPE} & \textbf{MAPE} \\
    \hline
   \textbf{ CNN-LSTM }&  3.68& 2.29\% & 0.18 & 1.04\% & \textbf{1.66\% }& 8.59 & 5.15\% & 0.45 & 2.20\% & \textbf{3.60\%} \\
    \hline
   \textbf{ LSTM-LSTM}& 4.36 & 2.71\% &  0.25 & 1.12\%  & \textbf{1.92\%} & 11.71 & 6.41\% & 0.51 & 2.59\% & \textbf{4.50\%} \\
    \hline
    \textbf{CNN-BiLSTM} & 2.81 & 1.76\% &  0.14 & 0.74\%  & \textbf{1.25\%} &  6.39& 3.90\% & 0.38 & 1.57\%  &  \textbf{2.77\%} \\
    \hline
  \textbf{LSTM-BiLSTM} & 0.82 & 0.61\% & 0.06 &0.35\%  & \textbf{0.47\%} & 2.72 & 1.70\% & 0.13 & 0.70\% & \textbf{1.20\%} \\
    \hline
    \end{tabularx}
    
    \end{center}
\end{table}

\begin{figure}[thpb]
      \centering
      
      \framebox{\parbox[c][13cm]{\textwidth}{
      
      \centering\includegraphics[scale=0.82]{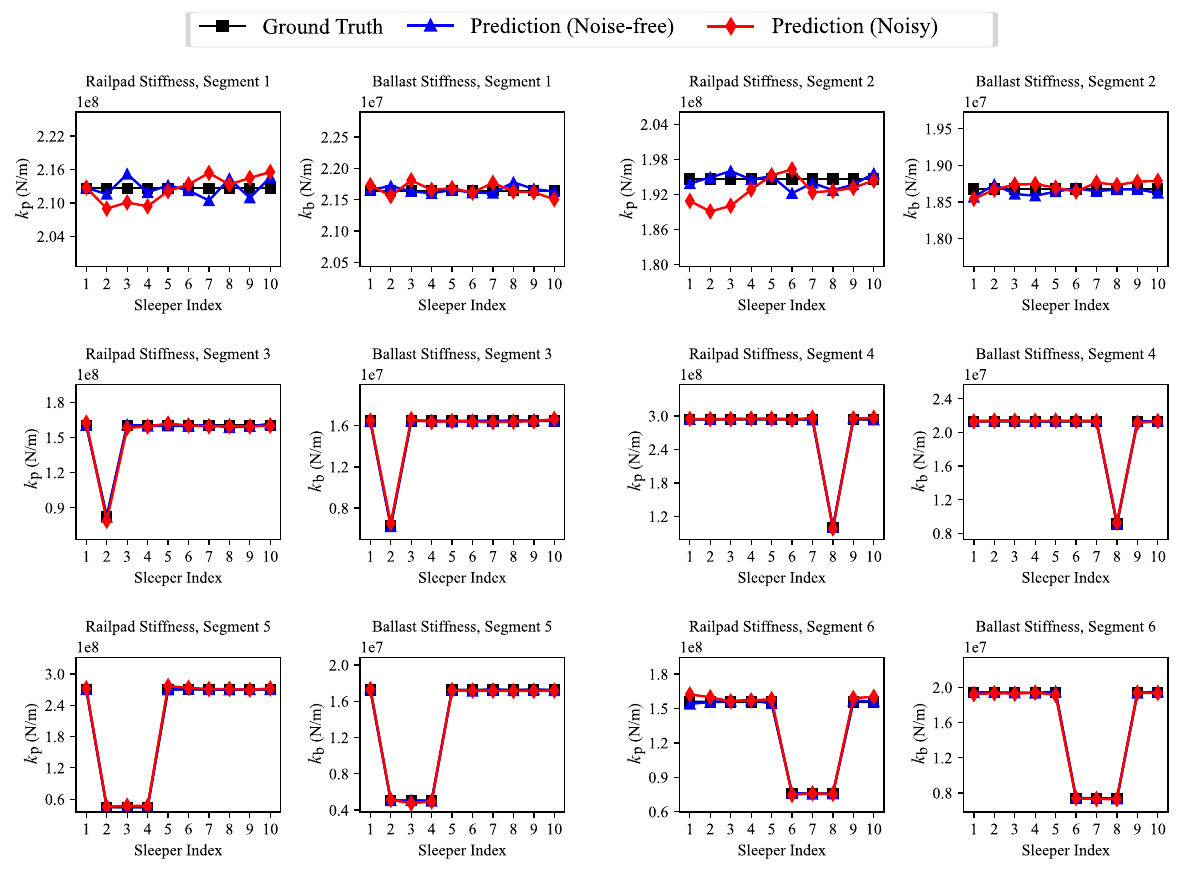}
      \vspace{-18pt}
}}
     
      \caption{LSTM-BiLSTM model's predictions on the noise-added test set, in the scenarios of constant and local changes of stiffness.}
      \label{predictions}
   \end{figure}

Figure \ref{predictions}  illustrates the ground truth and predictions for railpad and ballast stiffness in noise-free and noise-added conditions for six railway track segments. The ground truth values of stiffness parameters are represented by the black line, and the estimations in noise-free and noisy conditions are presented in blue and red lines, respectively. Railpad and ballast stiffness are estimated simultaneously in scenarios of constant track stiffness and local stiffness reduction in one and three sleepers.

\medskip

\subsection{Discussion}

The illustrative case study demonstrated that the proposed LSTM-BiLSTM model effectively estimates infrastructure stiffness using vehicle-based vibration signals, achieving a MAPE of 1.70\% for railpad stiffness and 0.70\% for ballast stiffness in the noise-added scenario. The BiLSTM networks outperformed their LSTM counterparts, highlighting the importance of incorporating bidirectional information to enhance infrastructure health monitoring algorithms. The analysis shows that employing BiLSTM in the higher-level health condition estimation phase can reduce MAPE and RMSE by almost 50\%, even in the presence of noise in the input signal. Hence, this model architecture holds promise for future research and practical applications, given the prevalence of various types of noise in drive-by vibration response measurements.

In the feature extraction phase, the results indicate a nuanced performance pattern among the four model architectures. The best-performing model, LSTM-BiLSTM, employs an LSTM layer in the feature extraction phase, demonstrating the potential of utilizing LSTM layers in this phase. This model outperforms the CNN-BiLSTM model, suggesting that LSTM's strength in capturing temporal dependencies can be more effective than CNNs when paired with BiLSTM networks. Moreover, the findings underscore the compatibility of BiLSTM networks with both CNN and LSTM feature extractors, as both CNN-BiLSTM and LSTM-BiLSTM models notably outperform their LSTM counterparts in noise-added scenarios.

The comparative performance of LSTM-LSTM and CNN-LSTM models introduces complexity in the evaluation of feature extraction methods. The LSTM-LSTM performance slightly falls behind the CNN-LSTM model, specifically in noise-added cases.  This necessitates further investigation into the utilization of LSTM networks in the feature extraction phase. The literature supports the significance of frequency domain analysis in feature extraction phases, as highlighted in studies such as \cite{shenEvaluatingRailwayTrack2023a, shenFastRobustIdentification2021, lamprea-pinedaRailwayTrackReceptance2024}. Both approaches, LSTM layers, and time-frequency transformations, focus on capturing temporal dependencies along the signal. LSTM networks achieve this through memory cells that learn patterns over time, making them promising candidates for feature extraction \cite{malhotra2016, houLSTMBasedAutoEncoderModel2020}. This similarity points to an area for future research, where further investigations could explore novel feature extraction methods, including LSTM layers, for infrastructure health monitoring models.

The employed CNN and LSTM feature extractors process vibration response data in the time domain. In contrast to previous studies that utilize time-frequency transformations of signals as input for the models \cite{lockeUsing2020, pathirageStructuralDamageIdentification2018a}, the proposed method eliminates the need for data preprocessing or handcrafted feature extraction. Instead, it provides an automatic health condition monitoring framework that learns relevant features directly through the training process. In this approach, we rely on the neural network’s model parameters to extract features from raw vibration signals. However, since drive-by vibration response signals are non-stationary, with components varying in both time and frequency, frequency decompositions may provide valuable insights over the signal \cite{nossierComparativeStudyTime2020}. Therefore, future research could investigate novel methodologies that integrate spectral information into deep learning frameworks. 

The results indicate that the estimation accuracy of all models decreases in the presence of noise. However, the proposed LSTM-BiLSTM model demonstrates greater robustness to noise compared to the other models, achieving the best performance in the noisy dataset. Despite this, its performance still declines when noise is introduced. Our models solely rely on the neural network parameters learning process to deal with the noise. However, recently, preprocessing techniques have been developed that employ denoising techniques to effectively deal with noisy conditions \cite{michauFullyLearnableDeep2022}. Therefore, future research could explore techniques to further enhance the model’s resilience to noise and to develop more robust deep-learning frameworks for noisy environments. This is especially important for real-world applications, where noisy data is a common challenge \cite{michauFullyLearnableDeep2022}. Moreover, the CNNs employed in the current paper stem from the Vanilla CNNs architecture employed in \cite{lockeUsing2020}. This can be compared with other machine learning models and recent CNN architectures like GoogLeNet and ResNet architectures \cite{szegedyInceptionv42017a}.

The proposed method effectively identifies and localizes multiple defects within the track segment, with floating locations. It simultaneously estimates railpad and ballast stiffness parameters and localizes stiffness reduction at a sleeper resolution. This enhances infrastructure monitoring tasks by providing detailed information on the type and severity of defects within the railway network. As a result, it can advance maintenance decision-making frameworks by offering comprehensive insights into the health conditions of the systems.

The illustrative case study considers a railway track infrastructure comprising beams and beam nodes. In this context, the rails and the sleepers function as the beams of the structure, while the locations of the sleepers are considered beam nodes that transfer the load to the ballast substructure. The proposed approach involves segmenting the vibration signal into bearing spans corresponding to individual sleepers or beam nodes. This methodology shows potential for application to other infrastructures with a similar configuration of beams and beam nodes, such as bridges. 

The dataset used in this study is derived from the vehicle-track interaction finite-element simulation proposed in \cite{shenEvaluatingRailwayTrack2023a}. While this case study demonstrates the proposed framework, simulated vehicle-track interaction signals may not fully capture all environmental and operational factors influencing drive-by measurements. For instance, the dataset in the current work assumes a constant measurement speed as the operational condition, whereas real-world drive-by measurements may involve speed variations. Additionally, the simulated dataset represents a single type of railway track structure, whereas real-world railway networks may exhibit structural variations. Therefore, this study opens new research directions for evaluating the performance of the proposed LSTM-BiLSTM network using field measurements of drive-by vibration responses.


\section{Conclusions and future work}

In this paper, we proposed a novel method for estimating and localizing multiple physical parameters of infrastructures using drive-by vibration responses, achieved through the development of an LSTM-BiLSTM network. Our methodology emphasizes the importance of considering both forward and backward temporal information through BiLSTM networks to enhance algorithms for monitoring infrastructure health conditions. The findings demonstrate that employing BiLSTM in the higher-level health condition estimation phase can significantly reduce MAPE and RMSE, even in the presence of noise. Moreover, the methodology highlights the potential of LSTM networks for extracting features from drive-by vibration signals. By accurately identifying the positions of vibration signals, our proposed framing approach improves the resolution of infrastructure monitoring to the beam level. Hence, the LSTM-BiLSTM model effectively estimates physical parameters at various components within the infrastructure beam levels, offering comprehensive insights into the health conditions of the systems and enhancing maintenance decision-making frameworks. In the case study analyzed, the results show that the method is effective in estimating the stiffness parameters of railpads and ballast, as well as in identifying track stiffness reductions at individual sleepers. The proposed BiLSTM networks significantly decrease the Mean Absolute Percentage Error (MAPE) and Root Mean Square Error (RMSE) by nearly 50\%, achieving a MAPE of 1.7\% for railpad stiffness and 0.7\% for ballast stiffness, even in noisy environments.
 
Future research will focus on three main research directions. First, the proposed hybrid model architecture requires further benchmarking against state-of-the-art methodologies. This includes considering advanced CNN and LSTM networks alongside other machine-learning methods. Second, future research should validate the performance of the proposed models based on field measurements and under real-world conditions. This further requires developing deep learning models that are robust to real-world noisy environments. Thirdly, the proposed framing approach requires high-accuracy signal localization, which can be challenging under certain operational conditions. In this regard, leveraging domain knowledge concerning the number of beam nodes in certain infrastructure lengths holds promise for addressing the signal localization challenge.

\section*{Author Contributions}
The authors confirm contribution to the paper as follows: study conception and design: R.R. Samani, A. Núñez, B. De Schutter; data collection: R.R. Samani, A. Alfredo Núñez; analysis and interpretation of results: R.R. Samani, A. Núñez; draft manuscript preparation: R.R. Samani, A. Núñez, B. De Schutter. All authors reviewed the results and approved the final version of the manuscript.

\section*{Declaration of Conflicting Interests}

The authors declared no potential conflicts of interest with respect to the research, authorship, and/or publication of this article.

\section*{Acknowledgments}

The first author would like to extend their gratitude to Chen Shen for providing the simulator used in this article. The first author would like to extend their gratitude to Yuanchen Zeng and Hesam Araghi for their valuable comments during the preparation of this article. This research was partly supported by ProRail and Europe’s Rail Flagship Project IAM4RAIL - Holistic and Integrated Asset Management for Europe’s Rail System [grant agreement 101101966].

\section*{Disclaimer}

Funded by the European Union. Views and opinion expressed are however those of the authors only and do not necessarily reflect those of the European Union. Neither the European Union nor the granting authority can be held responsible for them. This project has received funding from the European Union's Horizon Europe research and innovation programme under Grant Agreement No 101101966.

\vspace{0.1cm}

\begin{flushleft}
    \begin{minipage}{0.2\textwidth}
        \centering
        \includegraphics[width=\textwidth]{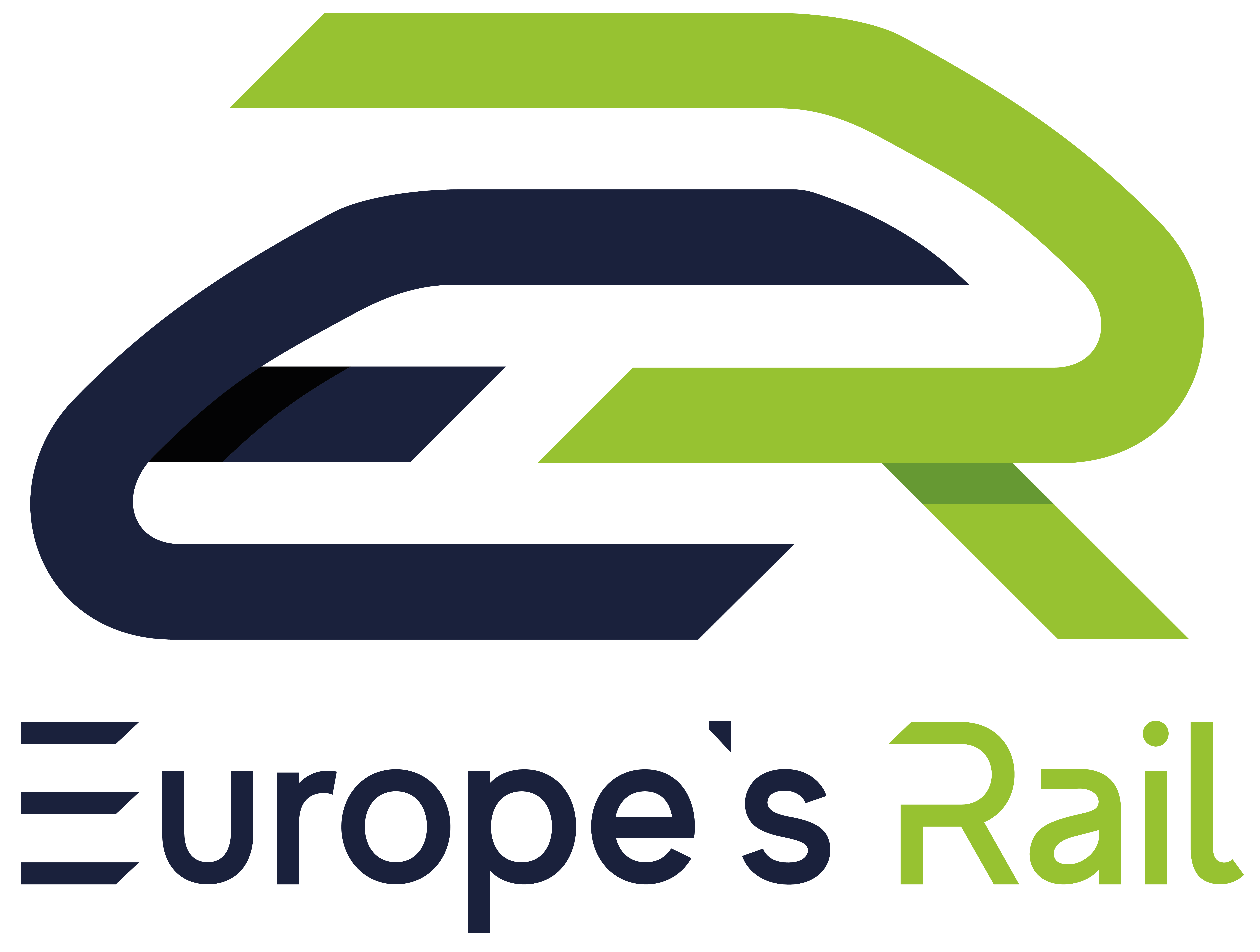} 
    \end{minipage}
    \hspace{1em} 
    \begin{minipage}{0.30\textwidth}
        \centering
        \includegraphics[width=\textwidth]{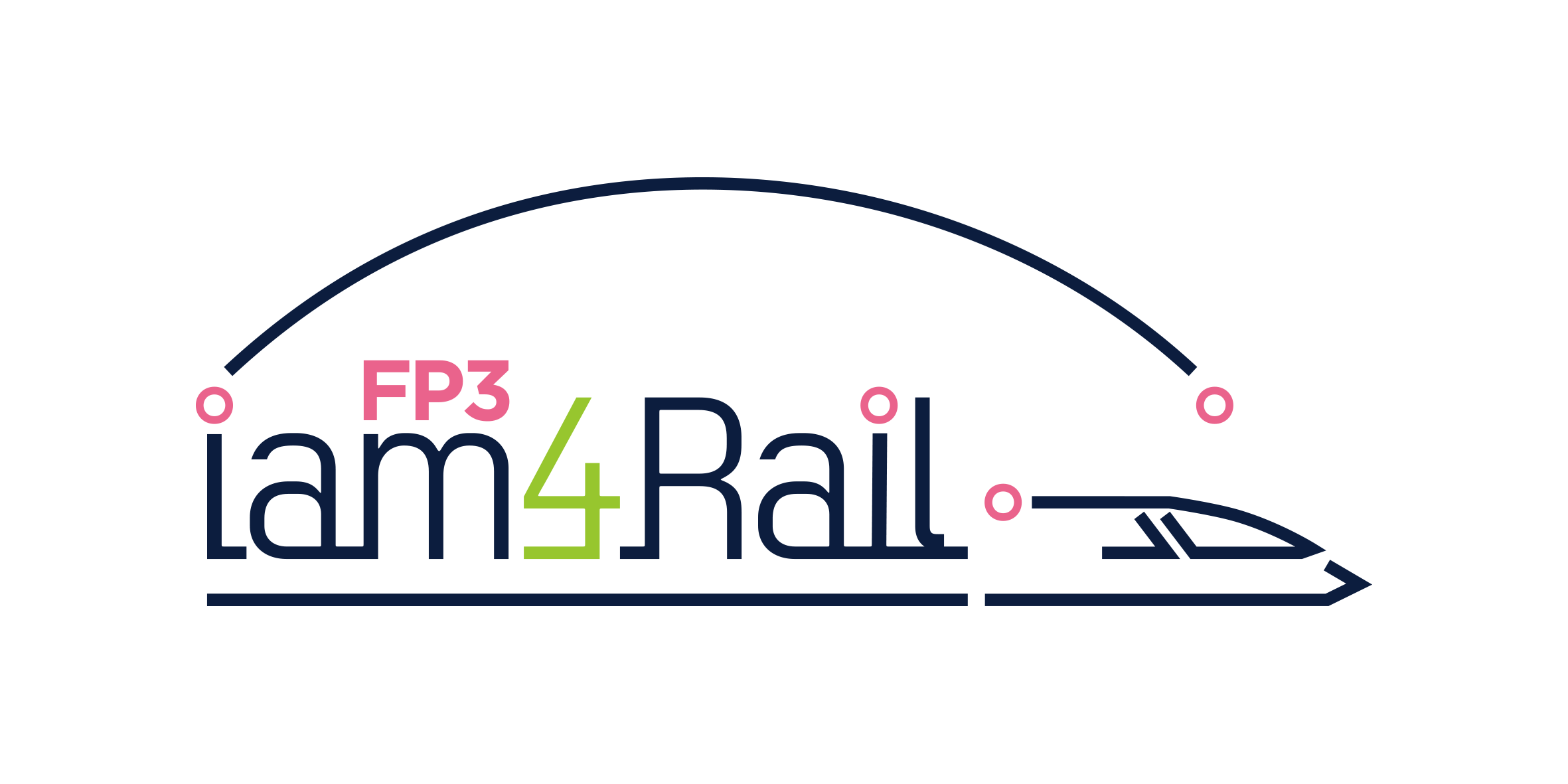} 
    \end{minipage}
\end{flushleft}

\newpage

\bibliographystyle{trb_09}
\bibliography{maintext}


\end{document}